*Article*

# Sensors for expert grip force profiling: towards benchmarking manual control of a robotic device for surgical tool movements

**Michel de Mathelin[1], Florent Nageotte[1], Philippe Zanne[1] and Birgitta Dresp-Langley[1,*]**

[1]ICube Lab, UMR 7357 CNRS, University of Strasbourg; demathelin@unistra.fr; nageotte@unistra.fr; zanne@unistra.fr; birgitta.dresp@unistra.fr;

* Correspondence: birgitta.dresp@unistra.fr;

**Abstract:** *STRAS* (*S*ingle access *T*ransluminal *R*obotic *A*ssistant for *S*urgeons) is a new robotic system based on the Anubis® platform of Karl Storz for application to intra-luminal surgical procedures. Pre-clinical testing of *STRAS* has recently permitted to demonstrate major advantages of the system in comparison with classic procedures. Benchmark methods permitting to establish objective criteria for 'expertise' need to be worked out now to effectively train surgeons on this new system in the near future. *STRAS* consists of three cable-driven sub-systems, one endoscope serving as guide, and two flexible instruments. The flexible instruments have three degrees of freedom and can be teleoperated by a single user via two specially designed master interfaces. In this study here, small force sensors sewn into a wearable glove to ergonomically fit the master handles of the robotic system were employed for monitoring the forces applied by an expert and a trainee (complete novice) during all the steps of surgical task execution in a simulator task (*4-step-pick-and-drop*). Analysis of grip-force profiles is performed sensor by sensor to bring to the fore specific differences in handgrip force profiles in specific sensor locations on anatomically relevant parts of the fingers and hand controlling the master/slave system.

1. Introduction

Flexible systems such as endoscopes are widely used for performing minimally invasive surgical interventions, as in intraluminal procedures or single port laparoscopy. Surgical platforms have been developed by companies and by laboratories to improve the capabilities of these flexible systems, for instance by providing additional Degrees of Freedom (DoF) to the instruments or triangulation configurations [1], [2]. In classic intraluminal procedures, the high number of DoF to be controlled represents a constraint, where several expert technicians, including the surgeon, have to work together in a complex environment. Robot assistance has been identified as a solution to this problem relative to the use of flexible systems in minimally invasive surgery [3], which explains the motivation for developing the new, teleoperated robotic system put to work in this study here. The goal of *STRAS* is to optimally assist the expert surgeon in minimally invasive procedures [4], and the design is based on the Anubis® platform developed by Karl Storz and the IRCAD [5]. Previous studies on *STRAS* were focused on the system architecture and the control theory of the application [4]–[6]. In minimally invasive surgical systems for endoscopic surgery, surgeons need to operate master interfaces to control the endoscope and surgical instruments. They need to be able to have optimal skills in controlling the system and the user interface for targeted manipulation of the remote-controlled slave system as well as to cope with the overall complexity of the design. Such expertise can only be achieved by learning to optimally master the control mechanisms through practice in a simulator task and in vivo. Human control of endoscopic surgical systems may benefit from robotic surgical assistance [7]. Previous studies were focused on tool-tip pressures and tactile feedback effects, rather than on the grip forces applied during manipulation of the handles [8]. The system described here was designed without force feedback, and maneuver control is therefore based solely on visual feedback from the 2D images provided by an endoscopic fisheye camera and displayed on a screen. Anthropometric data from the literature suggest that, with or without force feed-back, dynamic changes in perceptual hand and body schema representations and cognitive motor programming occur inevitably after repeated tool use [9], [10]. These cognitive changes reflect the processes which highly trained surgeons go through in order to adapt to the visual and tactile constraints of laparoscopic surgical interventions. Experts perform tool-mediated image-guided tasks significantly quicker than trainees, with significantly fewer tool movements, shorter tool trajectories, and fewer grasp attempts [11]. Also, an expert tends to focus attention mainly on target locations, while novices split their attention between trying to focus on the targets and, at the same time, trying to track the surgical tools. This reflects a common strategy for controlling goal-directed hand movements in non-trained operators in various goal-directed manual tasks [12], often considerably affecting task execution times. Such strategy variables are also likely to influence grip forces while manipulating the control sticks of a robotic device [13]. This work here is focused on the analysis of expertise and sensor specific force profiles during execution of a 4-step pick-and-drop task with the telemanipulation system of *STRAS*. Pre-clinical testing of the *STRAS* robotic system has permitted to demonstrate that an expert surgeon on his own can successfully perform all the steps of a complex endoscopic surgery task (colorectal endoscopic submucosal dissection) with the telemanipulation system [14-15]. Previously [16], we had shown that proficiency (expertise) in the control of the *STRAS* master/slave system is reflected by a lesser grip force during task execution as well as by a shorter task execution time. In the meantime, pre-clinical testing of the *STRAS* robotic system has permitted to demonstrate major advantages of the system for expert endoscopic surgeons in comparison with classic procedures [14-15], and benchmark measures permitting to establish objective criteria for expertise in using the system need to be found to ensure effective training of future surgeons on the system. Experimental studies of grip force strength and control for lifting and manipulating objects strategically have provided an overview of the contributions of each finger to overall grip strength and fine grip force control [17]. While the middle finger is the most important contributor to gross total grip force and, therefore, most important for getting a good grip of heavy objects to lift or carry, the ring finger and the small (pinky) finger are most important for the fine control of subtle grip force modulations [17], as those required for effectively manipulating the control handles of *STRAS*. Also, it is well-documented in the literature that grip force is systematically stronger in the dominant hand compared with the non-dominant hand [16, 18]. In this study here, the grip force profiles correspond to measurements collected from specific sensor positions on these anatomically relevant parts of finger and hand regions of the dominant and non-dominant hands. The grip force profiles of an expert in controlling the master/slave system are compared to those of an absolute beginner, who manipulated the robotic device for the first time. The wireless sensor glove hardware-software system described in [16], was improved and employed in this study here to collect force data from a novice trainee and an expert in various anatomical locations in the palm and on the phalanges of fingers of the right and left hands for detailed analyses in terms of sensor-specific grip force profiles.

## 2. Materials and Methods

*2.1 Slave Robotic System*

The slave robotic system is built on the Anubis® platform of Karl Storz. This system consists of three flexible, cable-driven sub-systems (for more information, [4]): one main endoscope and two lateral flexible instruments. The endoscope carries the camera providing the visual feedback at its tip, and has two lateral channels which are deviated from the main direction by two flaps at the distal extremity. The instruments have bending extremities (one direction) and can be inserted inside the channels of the endoscope. This system has a tree-like architecture and the motions of the endoscope act also upon the position and orientation of the instruments. Two kinds of instruments are available: electrical instruments and mechanical instruments. Overall, the slave system has 10 motorized DoF. The main endoscope can be bent in two orthogonal directions. This allows moving the endoscopic view respectively from left to right and from up to down, as well as forward / backward. Each instrument has three DoF: translation ($tz$) and rotation ($\theta z$) in the endoscope channel, and deflection of the active extremity (angle $\beta$). The deflection is actuated by cables running through the instrument body from the proximal part up to the distal end. The mechanical instruments can be opened and closed.

*2.2 Master/Slave Control*

The slave robot is controlled at the joint level by a position loop running at 1000 Hz on a central controller. The master side consists of two specially designed interfaces, which are passive mobile mechanical systems. The user grasps two handles, each having 3 DoF: they can translate for controlling instrument insertion, rotate around a horizontal axis for controlling instrument rotation, and rotate around a final axis (moving with the previous DoF) for controlling instrument bending. These DoFs are similar to the possible motions of the instruments as demonstrated in preclinical trials [15]. Each handle is also equipped with a trigger and with a small four-way joystick for controlling additional DoF. In the experiments here, the trigger is operated with the index finger of a given hand for controlling grasper opening and closing, the small joysticks for moving the endoscope are not used. Since there is no force measurement on the slave side, no force effects are reproduced on the master side.

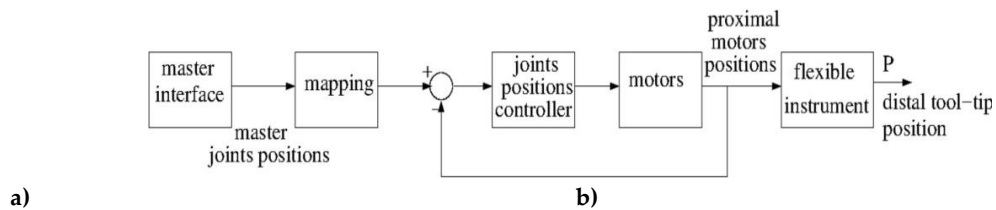

a)          b)

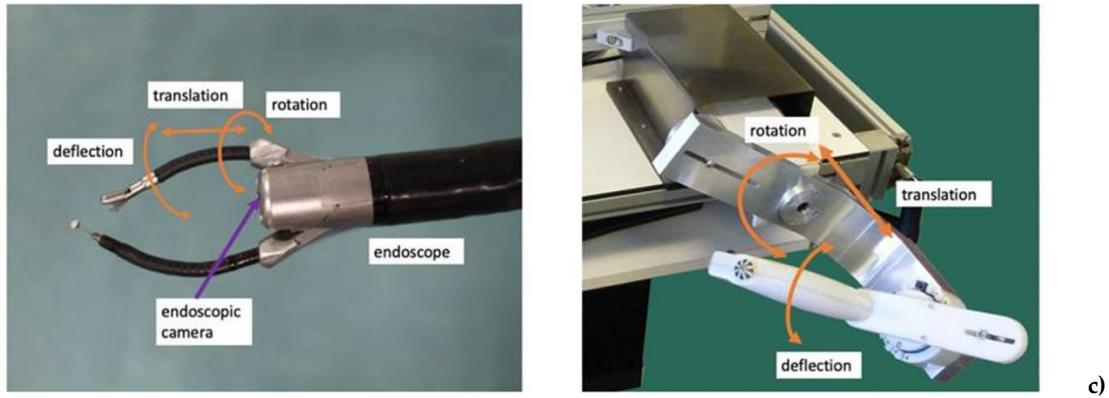

**Figure 1.** Expert wearing the sensor gloves while manipulating the handles of the robotic master/slave system (**a**). Master-slave control chart of the system (**b**). Direction and type of tool-tip and control movements **(c).**

A high-level controller running on a computer under a real-time Linux OS communicates with the master interfaces and provides reference joint positions to the slave central controller. The user sits in front of the master console and looks at the endoscopic camera view displayed on the screen in front of him/her at a distance of about 80 cm while holding the two master handles, which are about 50 cm away from each other. Seat and screen heights are adjustable to optimal individual comfort. The two master interfaces are identical and the two slave instruments they control are also identical. Therefore, for a given task the same movements need to be produced by the user whatever the hand he/she uses (left or right). The master interfaces are statically balanced and all joints exhibit low friction, and therefore only minimal forces are required to produce movements in any direction. A snapshot view of a user wearing the sensor gloves while manipulating the handles of the system is shown in Figure 1a here above. The master-slave control chart of the master/slave system is displayed in Figure 1b. Figure 1c shows the different directions and types of tool-tip and control movements.

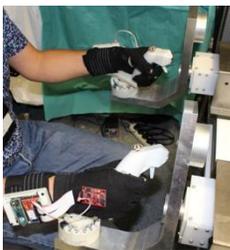

### 2.3 Glove Design for Grip Force Profiling

STRAS has its own grip style design and, therefore, specific gloves, one for each hand, with inbuilt Force Sensitive Resistors (FSRs) were developed to measure the two male individuals' left and right hand grip forces applied to the two handles of *STRAS* for controlling and operating the master/slave system. The hardware and software configurations are described here below.

*2.3.1 Hardware*

The gloves designed for the study contain 12 FSR, in contact with specific locations on the inner surface of the hand as given in Figure 2. Two layers of cloth were used and the FSRs were inserted between the layers. The FSRs did not interact, neither directly with the skin of the subject, nor with the master handles, which provided a comfortable feel when manipulating the system. FSRs were sewn into the glove with a needle and thread. Each FSR was sewn to the cloth around the conducting surfaces (active areas). The electrical connections of the sensors were individually routed to the dorsal side of the hand and brought to a soft ribbon cable, connected to a small and very light electrical casing, strapped onto the upper part of the forearm and equipped with an Arduino microcontroller. Eight of the FSR, positioned in the palm of the hand and on the finger tips, had a 10 mm diameter, while the remaining four located on middle phalanxes had a 5mm diameter. Each FSR was soldered to 10KΩ pull-down resistors to create a voltage divider, and the voltage read by the analog input of the Arduino is given by (1)

$$V_{out} = R_{PD}V_{3.3}/(R_{PD}+R_{FSR}) \tag{1}$$

where $R_{PD}$ is the resistance of the pull down resistor, $R_{FSR}$ is the FSR resistance, and $V_{3.3}$ is the 3.3 V supply voltage. FSR resistances can vary from 250Ω when subject to 20 Newton (N) to more than 10MΩ when no force is applied at all. The generated voltage varies monotonically between 0 and 3.22 Volt, as a function of the force applied, which is assumed uniform on the sensor surface. In the experiments here, forces applied did not exceed 10N, and voltages varied within the range of [0; 1500] mV. The relation between force and voltage is almost linear within this range. It was ensured that all sensors provided similar calibration curves. Thus, all following comparisons are directly between voltage levels at the millivolt scale. Regulated 3.3V was provided to the sensors from the Arduino. Power was provided by a 4.2V Li-Po battery enabling use of the glove system without any cable connections. The battery voltage level was controlled during the whole duration of the experiments by the Arduino and displayed continuously via the user interface. The glove system was connected to a computer for data storage via Bluetooth enabled wireless communication running 115200 bits-per-second (bps).

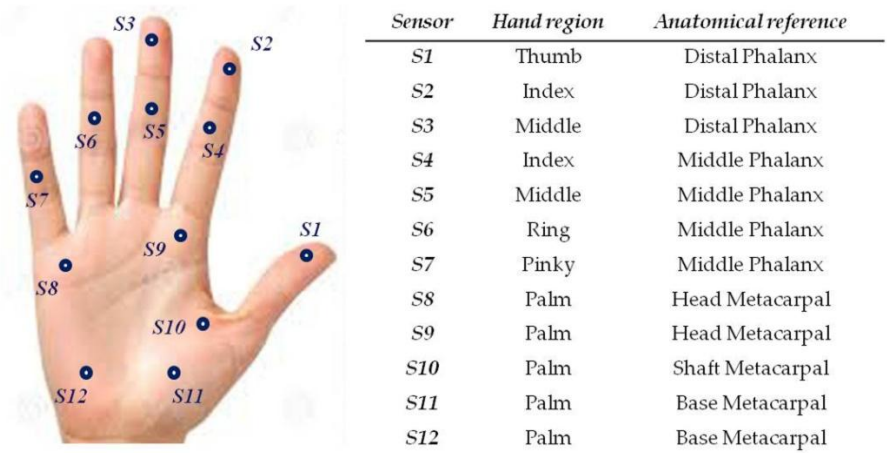

**Figure 2.** Sensor locations on the inner surface of the hand

*2.3.2 Software*

The software of the glove system was divided into two parts: one running on the gloves, and one running on the computer algorithm for data collection. The general design of the glove system is described as follows. Each of the two gloves was sending data to the computer separately, and the software read the input values and stored them on the computer according to their header values indicating their origin. The software running on the Arduino was designed to acquire analog voltages provided by the FSR every 20 milliseconds (50Hz). In every loop, input voltages were merged with their time stamps and sensor identification. This data package was sent to the computer via Bluetooth, which was decoded by the computer software. The voltages were saved in a text file for each sensor, with their time stamps and identifications. Furthermore, the computer software monitored the voltage values received from the gloves via a user interface showing the battery level. In case the battery level drops below 3.7 V, the system warns the user to change or charge the battery. Such an event did not occur during the experiments reported here. Figure 3 here below shows a snapshot view of the right-hand glove in action (a) and the general design chart of hard-to-software operating system (b).

*2.4 Experimental Task Design*

For this user grip force profiling study, a *4-step pick-and-drop task* was designed. Four snapshot views of the four task steps are shown in Figure 3. During the experiments, only one of the two instruments controlling the tool-tips (left or right, depending on the task session) was moved, while the main endoscope and its image remained still. The experiments started with the right or left hand gripper being pulled back. Then the user had to approach the object (*step 1*) with the distal tool extremity by manipulating the handles of the master system effectively.

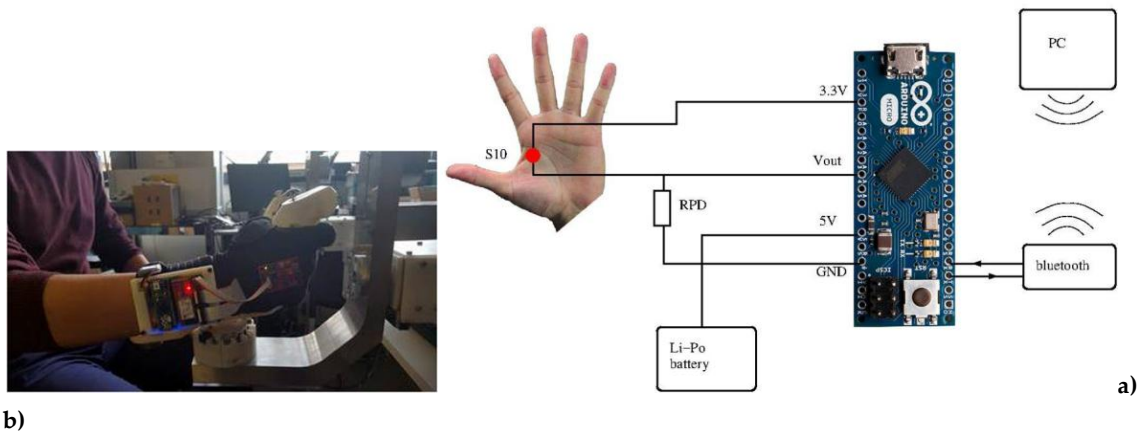

**Figure 3.** Snapshot view of right-hand glove in action (**a**). Design chart of the single-sensor-to-software operating system (**b**).

Then, the object had to be grasped with the tool (*step 2*). Once firmly held by the gripper, the object had to be moved to a position on top of the target box (*step 3*) with the distal extremity of the tool in the correct position for dropping the object into the target box without missing (*step 4*). To drop the object, the user had to open the gripper of the tool. The user started and ended a given task session by pushing a button, wirelessly connected to the computer. One expert, who had been practicing with the system since its manufacturing and who is currently the most proficient user, and one complete novice who had never used the system and had no prior experience with any similar surgical system participated in the grip force profiling experiments.

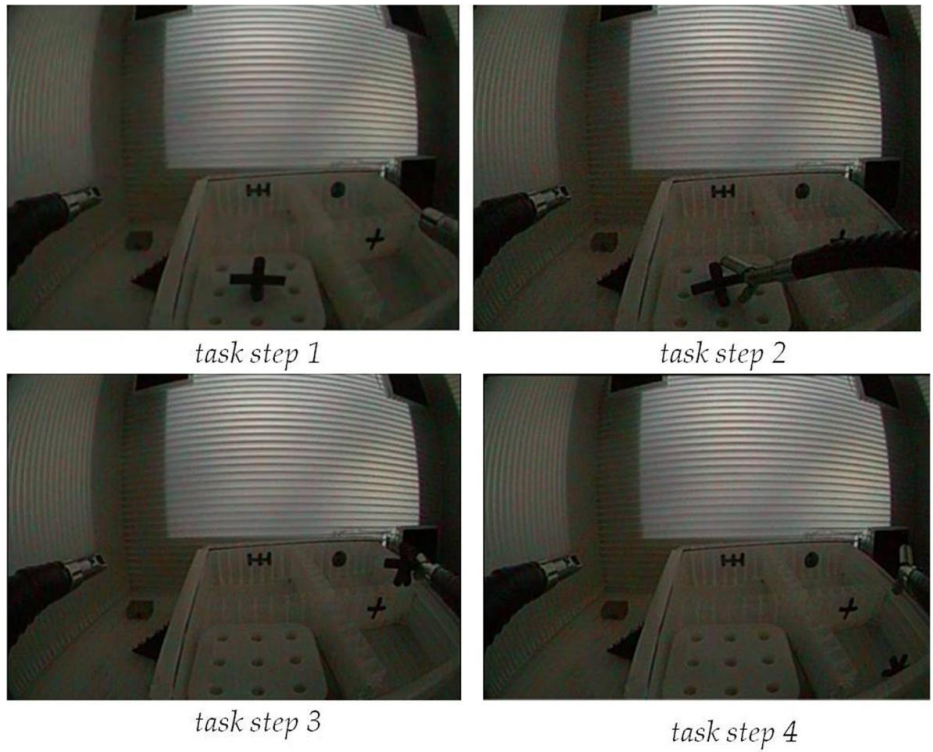

**Figure 4.** Snapshot views of the four successive steps of the pick-and-drop task when executed with the right hand by manipulating the corresponding instrument of the robotic system.

The expert and the novice's hand sizes were about the same, and the sensor gloves were developed specifically to fit the hands of these two individuals. The expert user was left handed and the novice user was right handed. As explained earlier, the left and right interfaces are identical, and the same task is realized with both hands. Between nine and eleven consecutive sessions were recorded for the two subjects, for task execution with their right and left hands. Before the experiment started, the novice user was made familiar with the buttons and the running of the system. Force data were collected from all twelve sensor locations for both individuals and both hands, left and right.

## 3. Results

In a first preliminary analysis, the data from all sensors from left-hand and right-hand task execution were plotted as a function of the individual total number of grip force measures in millivolt sampled in time across all individual sessions. The function that links the sensor output voltage to force is almost linear within the measured range, as shown here below in Figure 5.

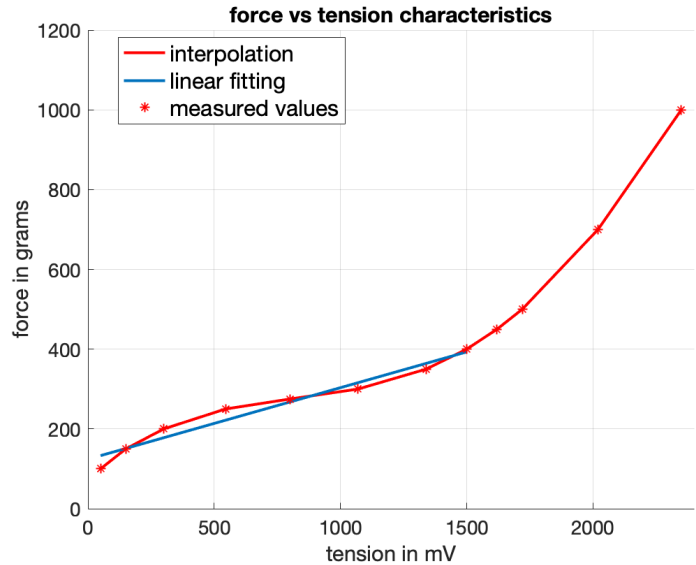

**Figure 5.** Static force in grams as a function of the tension output in mV of the sensors. The relation is almost linear between 50 mV and 1500mV.

The total number of sessions differs slightly between the two subjects and between hand conditions. The results from these preliminary, descriptive analyses in terms of means and standard deviations of individual grip force data for task execution with the dominant or non-dominant hand, collected across sessions, are shown in Table 1. It is shown that the sensors S5, S6, S7 and S10 produced reliable and consistent output values across sessions and hand conditions for the expert and the novice allowing for a statistical comparison between their individual grip force profiles. The means and standard deviations shown here for the expert were computed on the basis of a total of 5117 grip force data for the non-dominant right hand recorded in 12 successive sessions and a total of 4442 grip force data for the dominant left hand recorded in 10 successive sessions. The statistics for the novice were computed on the basis of a total of 6497 grip force data for the non-dominant left hand recorded from 10 successive sessions, and a total of 8483 data for the dominant right hand, recorded from 11 successive task sessions.

Sensors S5, S6, S7, and S10 were positioned on regions of the fingers and the hand that are particularly critical for general grip force and/or subtle grip force control during task execution. Sensor S5 was positioned on the middle phalanx of the middle finger, critical for strong grip force control, sensor S6 on the middle phalanx of the ring finger, contributes to force modulation less critical for strong grip control, sensor S7 on the middle phalanx of the small finger (pinky) highly critical for subtle, finely tuned grip force control. Experts like surgeons use their pinkies strategically for the fine-tuning of hand-tool interactions [17, 19, 20]. Sensor S10 was positioned on the metacarpal that joins the thumb to the wrist, important in general grip control (getting hold of the device handles), but not critical for subtle strategic grip force control [17].

**Table 1.** Means and standard deviations of the expert's and the novice's individual grip force data (in millivolts (mV)) collected from each sensor during task execution in successive sessions with the dominant and non-dominant hand.

*Dominant hand*

|  | S1 | S2 | S3 | S4 | S5 | S6 | S7 | S8 | S9 | S10 | S11 | S12 | Sensor |
|---|---|---|---|---|---|---|---|---|---|---|---|---|---|
| *Expert* | 0 | 1.4 | 4.5 | 2 | 99 | 452 | 587 | 0 | 0.5 | 474 | 0 | 1.2 | Mean |
|  | 0 | 0.7 | 1.6 | 1.2 | 89 | 102 | 53 | 0 | 7.7 | 70 | 0 | 1.7 | Std.dev. |
| *Novice* | 0 | 23 | 674 | 0.7 | 754 | 498 | 85 | 651 | 1132 | 617 | 847 | 858 | Mean |
|  | 0 | 150 | 207 | 5.5 | 188 | 74 | 49 | 192 | 483 | 312 | 418 | 280 | Std.dev. |

*Non-dominant hand*

|  | S1 | S2 | S3 | S4 | S5 | S6 | S7 | S8 | S9 | S10 | S11 | S12 | Sensor |
|---|---|---|---|---|---|---|---|---|---|---|---|---|---|
| *Expert* | 1 | 0 | 0 | 9 | 364 | 371 | 71 | 109 | 90 | 160 | 825 | 418 | Mean |
|  | 1.5 | 0 | 0 | 27 | 107 | 68 | 37 | 118 | 170 | 138 | 450 | 250 | Std.dev. |
| *Novice* | 0 | 69 | 0 | 0 | 296 | 1063 | 526 | 233 | 0 | 500 | 0 | 0.4 | Mean |
|  | 0 | 27 | 0 | 0 | 148 | 120 | 64 | 257 | 2 | 365 | 0 | 0.5 | Std.dev. |

The sensors, S1, S2, S3 and S4 were all placed on distal phalanxes, which were not needed for producing task-critical tool-movements here in this task. The index finger is minimally needed for triggering movements relative to opening and closing the grippers of the system. S8, S9, S11 and S12 produced markedly different grip force data across the two individuals in all the sessions and across hand conditions. These differences in grip-force profiles are explained by the fact that the corresponding finger or hand regions were not used in the same way, for either general grip force control or strategic grip force deployment in the manipulation of the robotic system, by the expert and the novice, who was an absolute beginner ad had no experience at all with the system. These findings show promising differences that could be exploited in future studies on larger study populations, where the expert grip force profiles for strategically selected sensor locations may serve as benchmarks for assessing the skill status or evolution of novices and/or absolute beginners from different sample populations. For now, the data here will be exploited in further statistical analyses that are to highlight some of the key aspects of the differences in the grip-force profiles between an expert user and an absolute beginner. For now, data from sensor locations that produced zero-signal profiles in any (one or more) of the conditions tested were not taken into account for further statistical comparisons. As a consequence, the analyses shown and discussed here below are selective to sensor locations S5, S6, S7 and S10. They are to serve as examples, and to provide guidance for further benchmarking in future studies on larger sample populations.

*3.1. Dynamic grip force range in the expert and novice data at selected sensor locations*

To gain a descriptive overview of the dynamic range (upper and lower limits) of the grip force distributions from each of the four relevant sensor positions S5, S6, S7 and S10 during task execution with the dominant and non-dominant hands across sessions, these sensor data were represented in terms of box-plots for each subject (expert and novice) and for each of their hands (dominant and non-dominant). These box-plots are shown here below in Figure 6.

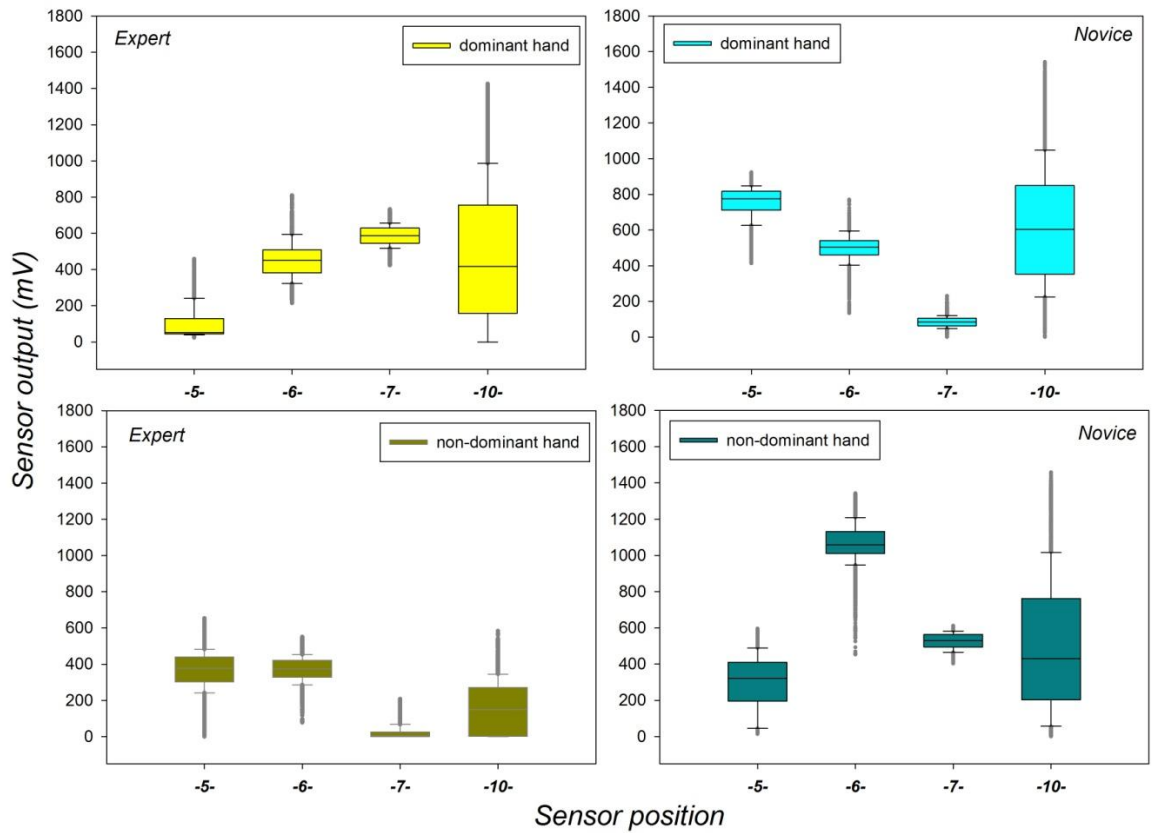

**Figure 6.** Dynamic range of grip force data recorded for the expert and the novice from each task- relevant sensor position in successive sessions with the dominant and non-dominant hands

The box-plots reveal marked differences in upper and lower limits of the grip force distributions from the different sensors. The output range is found to vary as a function of expertise, sensor position and handedness. The largest amount of grip force variability is found in sensor S10, positioned on the metacarpal that joins the thumb to the wrist, important in general grip control for getting a good hold of the device handles, but not critical for subtle strategic grip force control for finely tuned task maneuvers [17]. The smallest amount of grip force variability is found in sensor S7, positioned on the middle phalanx of the small finger (pinky) and highly critical for subtle grip force control during task maneuvers. The most noticeable differences between the grip force distributions of the expert and the novice are observed in sensor positions S7 and S5 on the middle finger, critical for strong grip force modulation and control. The total amount of grip force deployed on a given sensor is shown to depend on expertise, and on whether the dominant or the non-dominant was used to manipulate the robotic system. These observations suggest complex interactions between factors warranting a series of detailed analyses of variance (ANOVA). The design and outcome of these statistical analyses are described in the following subsections.

*3.2. Individual sensor-specific effects of task session and handedness with interactions*

In the next step, we analyzed the individual grip force profiles of the expert and the novice for task execution with each hand, the dominant and the non-dominant one, across the successive individual sessions for each of the four relevant sensors S5, S6, S7 and S10. The data from this analysis are shown in Figure 7.

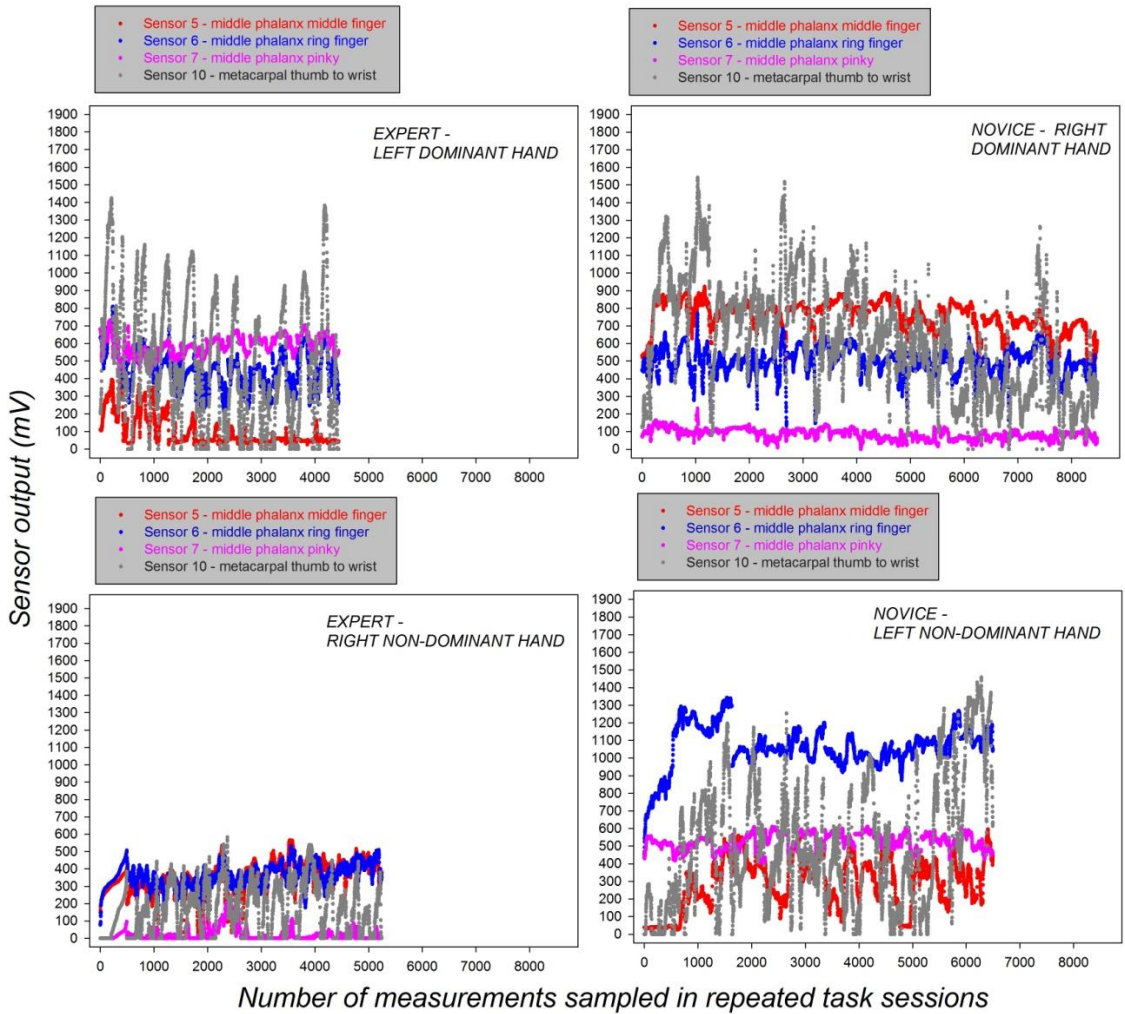

**Figure 7.** Grip force profiles of the expert (left) and the novice (right) from the relevant sensors for task execution with the dominant and the non-dominant hand across successive individual sessions.

The data plotted in Figure 7 reveal distinct grip force profiles of the novice and the expert, depending on whether they use their dominant or their non-dominant hands. The grip force profiles of the novice systematically display stronger grip forces in the non-dominant hand compared with those of the expert's non-dominant hand irrespective of sensor position. A marked dependency of individual grip forces on sensor position is seen when comparing the grip force profiles of the two subjects using their dominant hands. While the expert's dominant hand systematically displays noticeably stronger grip forces compared with the novice's in the profiles for sensor S7 positioned on the pinky finger, exactly the reverse is observed for sensor position S5 on the middle finger, where the expert's dominant hand displays minimal grip force strength, while the profile of the novice displays forces up to eight times the strength of those of the expert, especially towards the end of the successive task sessions.

To assess the statistical significance of these effects, analyses of variance on the raw data were performed. In each of the four relevant sensors S5, S6, S7 and S10, and in each subject, the expert and the novice, we tested for statistically significant effects of task repetition (training), reflected by the factor task session, and of handedness on the individual grip force data. To this effect, the individual raw data of each subject were submitted to several two-way analyses of variance using the general linear model for $Subject_1 x Sensor_1 x Hand_2 x Session_{10}$, with one level of the 'subject' factor, two levels of the *Hand* factor ('dominant' *versus* 'non-dominant'), and ten levels of the *Session* factor. Given that we have a total of 10 sessions for the dominant and a total of 11 sessions for the non-dominant hand in the case of the expert, and a total of 11 sessions for the dominant hand, and a total of 10 sessions for the non-dominant hand in the case of the novice, the analyses of variance were performed on the first ten successive sessions for each hand in each subject. This Cartesian analysis plan enabled the computation of interactions between the *Session* and *Hand* factors for each subject and sensor. Grip force data in terms of means ($M_{1-10}$) and their standard errors (*SEM*) from sensor-specific individual two-way analyses of variance for ten successive individual task sessions are summarized in TableS1 in the Supplementary Material Section.

**Table 2.** F-statistics and probability limits relative to the effects of the *Hand* and *Session* factors and their interactions from the two-way analyses of variance for each subject and sensor.

### *Expert*

|  | S5 | S6 | S7 | S10 |
|---|---|---|---|---|
| *Hand* | $F(1,9685)=3017; p<.001$ | $F(1,9685)=3734; p<.001$ | $F(1,9685)=5745; p<.001$ | $F(1,9685)=3339; p<.001$ |
| *Session* | $F(9,9685)=213; p<.001$ | $F(9,9685)=209; p<.001$ | $F(9,9685)=307; p<.001$ | $F(9,9685)=47; p<.001$ |
| *HandxSession* | $F(9,9685)=747; p<.001$ | $F(9,9685)=277; p<.001$ | $F(9,9685)=295; p<.001$ | $F(9,9685)=138; p<.001$ |

### *Novice*

|  | S5 | S6 | S7 | S10 |
|---|---|---|---|---|
| *Hand* | $F(1,13613)=8655; p<.001$ | $F(1,13613)=1499; p<.001$ | $F(1,13613)=6755; p<.001$ | $F(1,13613)=200; p<.001$ |
| *Session* | $F(9,13613)=413; p<.001$ | $F(9,13613)=102; p<.001$ | $F(9,13613)=188; p<.001$ | $F(9,13613)=201; p<.001$ |
| *HandxSession* | $F(9,13613)=677; p<.001$ | $F(9,13613)=152; p<.001$ | $F(9,13613)=359; p<.001$ | $F(9,13613)=1073; p<.001$ |

The F-statistics relative to the effects of the *Hand* and *Session* factors and their interactions, with their corresponding probability limits ($p$), are shown in Table 2 here above. Since the effect sizes in terms of differences between means are difficult to grasp from looking at the tables, we represented the average results graphically in Figure 8 here below, which also permits highlighting interactions visually.

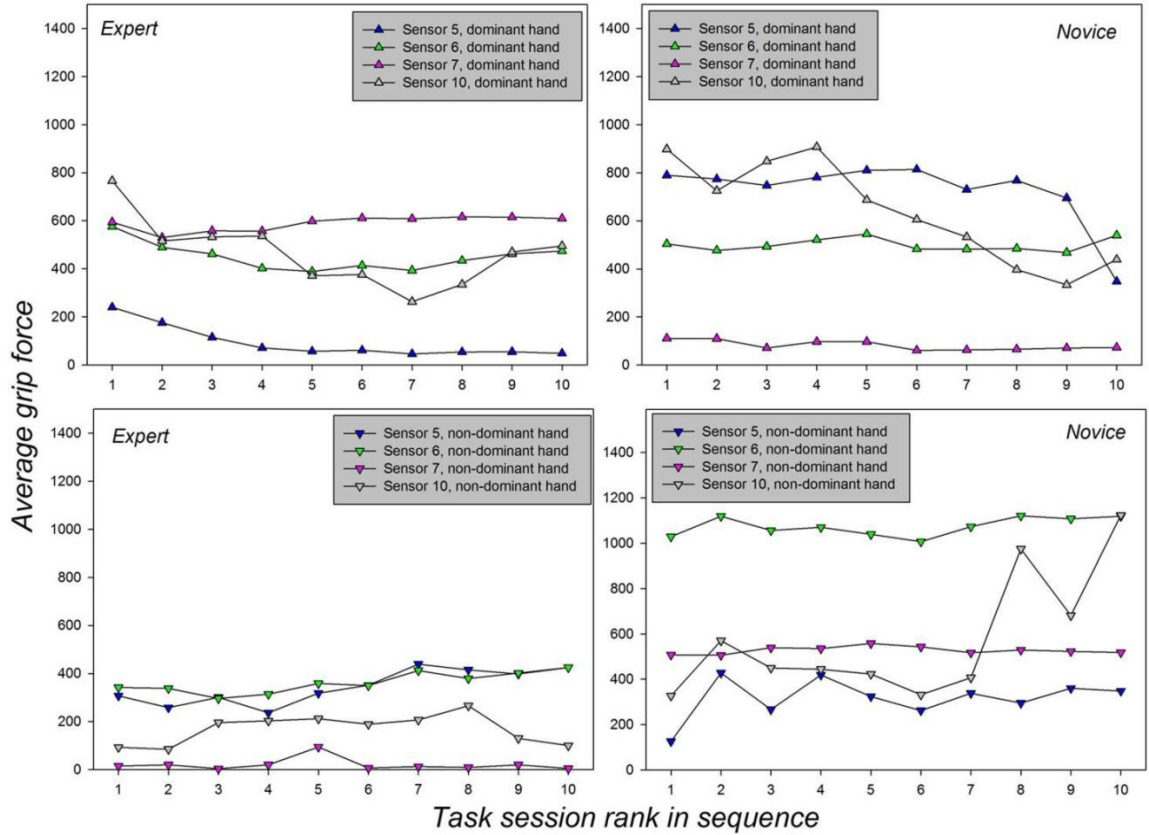

**Figure 8.** Average grip forces, reflected by sensor output in mV, are plotted as a function of the session and hand factor for the expert (left) and the novice (right) performing the task with their dominant (top) and non-dominant (bottom) hands.

The effect sizes from the two-way analyses of variance, shown graphically in Figure 8 and reflected by differences in average grip forces, reveal significant trends towards a decrease in the dominant hand as the sessions progress, and towards an increase in the non-dominant hand as the sessions progress, especially for sensor 10 on the metacarpal between thumb and wrist of the novice, presumably as a result of task fatigue: the novice presses harder with the wrist to compensate for lack of fine grip force control. The statistically significant effect of handedness expresses itself differently in the two subjects; while the expert mostly deploys stronger average grip force with his dominant hand, at most sensor positions, the novice deploys stronger average grip force with his non-dominant hand at most sensor positions. While the force profiles from sensor 7 of the expert's dominant hand indicate proficient use of the pinky finger for subtle grip force control with the dominant hand, the weak average grip forces from sensor 7 in his non-dominant hand, resembling those of the novice's dominant hand, reflect the expert's lack of fine grip force control in the non-dominant hand: the expert is, indeed, not trained in performing the task with his non-dominant hand. Expertise, or proficiency, is reflected in the individual grip force profiles by a strategic and consistently parsimonious deployment of grip forces in the hand that has been trained for controlling the robotic device. The novice, on the other hand, deploys forces non-strategically, in the dominant and the non-dominant hand. Significant interactions between the *Hand* factor and the *Session* factor' are found in each of the two subjects, and in each of the four sensors. This leads to conclude on a complex joint influence of handedness and repeated training on an individual's grip force profiles. Interactions between handedness and repeated training are, as would be expected, partly independent of an individual's proficiency level, i.e. whether he is an expert at performing the given task or not, as even expert performance is subject to variations and likely to evolve.

*3.3. Effects of expertise as a function of sensor position*

The grip force profiles from the hand of the expert trained in the specific task given, which is generally the dominant hand, allow to bring to the fore specific characteristics of expert performance compared with that of an absolute beginner, i.e. a novice performing the same task under the same conditions for the first time. The next step of the analysis is aimed at further highlighting critical strategy relevant differences in the grip force profiles, comparing the expert performing the task with his dominant hand to the novice performing the task with his dominant hand. To test for significant differences between subjects (proficiency level) and sensors, and for significant interactions between proficiency level and sensor location, the raw force data of the expert and the absolute novice performing the task with their dominant hands only, in all sessions, were considered. The data from both subjects, all four sensors, and all sessions, were submitted to a single two-way analysis of variance using the general linear model for the analysis plan *Subject$_2$xSensor$_4$*, with two levels of the *Subject* factor (*expertise* factor) and four levels of the *Sensor* factor. The F-statistics relative to the effects of the *Subject* and *Sensor* factors and their interactions, with their corresponding probability limits (*p*), are shown in Table 3 Statistics from the post-hoc comparisons for effects of expertise within each sensor are shown in Table 4.

**Table 3.** F-statistics from the two-way analysis of variance for effects of expertise and sensor.

*Subject (expertise)*   $F(1,51692)=2680; p<.001$

*Sensor*   $F(3,51692)=2840; p<.001$

*Subject x Sensor*   $F(3,51692)=2569; p<.001$

The results from this analysis show a statistically significant effect of the *Subject* (*expertise*) factor, a statistically significant effect of the *Sensor* factor, and a statistically significant interaction between these two factors. This leads to conclude that the expert and the absolute novice do not use the different anatomical locations on which the sensors were placed in the same way, but employ significantly different grip force control strategies, reflected by the statistically significant differences in the sensor output data. These effects and their interaction are shown graphically in Figure 9, where individual force data of the expert and the absolute novice were plotted for each sensor and individual task session in time.

**Table 4.** Statistics and probability limits from the Holm-Sidak *post-hoc* comparisons for effects of expertise within each sensor

*Sensor 5*

| | Diff of Means | t | Unadjusted P | Critical Level |
|---|---|---|---|---|

| | Diff of Means | t | Unadjusted P | Critical Level |
|---|---|---|---|---|
| *Expert vs Novice* | 655.653 | 198.175 | <.001 | .050 |

*Sensor 6*

| | Diff of Means | t | Unadjusted P | Critical Level |
|---|---|---|---|---|
| *Expert vs Novice* | 46.167 | 13.954 | <.001 | .050 |

*Sensor 7*

| | Diff of Means | t | Unadjusted P | Critical Level |
|---|---|---|---|---|
| *Expert vs Novice* | 502.306 | 151.825 | <.001 | .050 |

*Sensor 10*

| | Diff of Means | t | Unadjusted P | Critical Level |
|---|---|---|---|---|
| *Expert vs Novice* | 14.069 | 43.24 | <.001 | .050 |

The graphs in Figure 9 here below display distinct grip force profile telling apart the control strategies of the expert and the novice. The middle finger, preferentially used for controlling subtle grip force modulations in surgery and other manual tasks, is considerably solicited by the novice in all sessions, significantly less by the expert, as clearly shown here on the basis of the individual grip force profiles for sensor location 5, on the middle phalanx of the middle finger (top left graph). The small finger (pinky), strategically used by surgeons and other experts in various manual tasks for strong grip force control, is consistently solicited by the expert and significantly less by the novice, who almost applied no force at all to that finger region, as shown here on the basis of the individual grip force profiles for sensor location 7, on the middle phalanx of the pinky finger (bottom left graph). The ring finger contributes largely to grip force control, much less to total force magnitude than the middle and small fingers. The individual grip force profiles for the corresponding sensor location 8 show that the expert applies force to that finger region in a different manner compared with the novice (top right graph). Although the differences between the two subjects here may appear small in the graph, they are systematic and statistically significant in the light of *2x2 post-hoc comparisons* shown in Table 4 here above. The *post-hoc* tests were performed using the Holm-Sidak method. Similar conclusions are drawn from the individual grip force profiles for the sensor location 10, on the metacarpal joining the thumb to the wrist. This hand region is exploited for maintaining constant grip force on the handles of the robotic device, and the fully proficient expert needs to apply less force to achieve that goal than a novice, as consistently shown here.

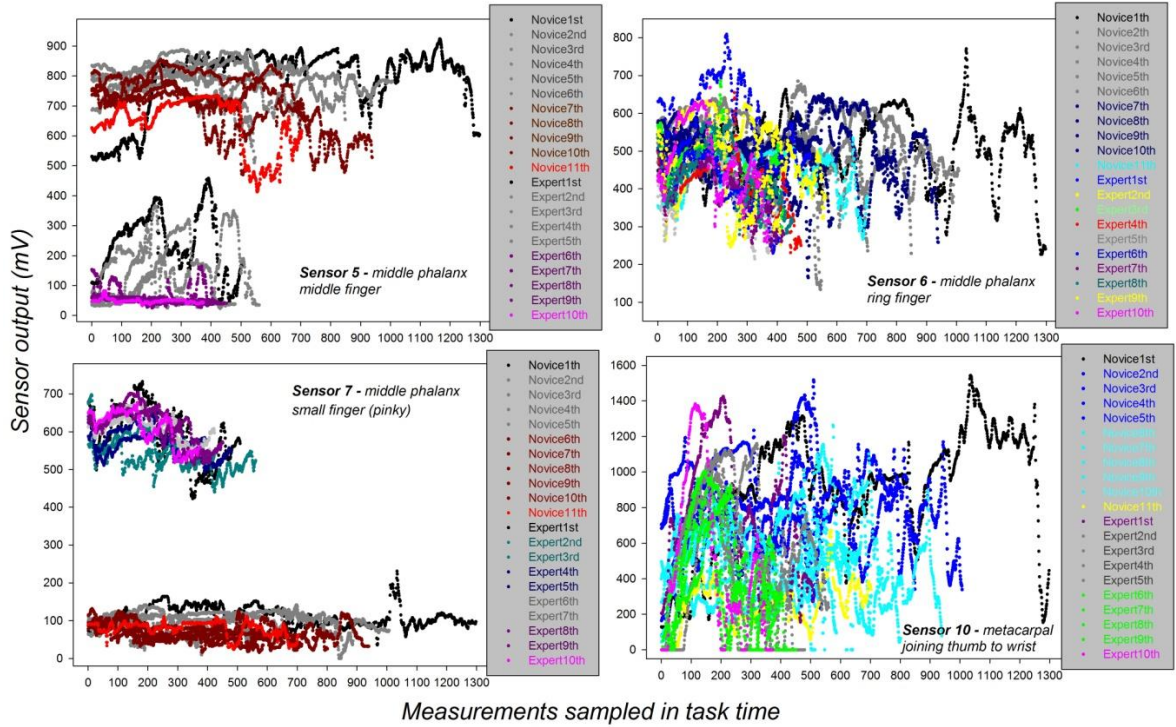

**Figure 9.** Distinct grip force profiles, plotted separately for each of the four specific sensor locations, from the successive sessions of the expert and the novice performing the *pick-and-drop* task on the robotic system with their dominant hands.

The next analysis was aimed at further highlighting subtle ways in which the interaction between task proficiency (expertise) and sensor location expresses itself at the beginning and at the end of the repeated task sessions. To that effect, the individual force data of the expert and the novice were plotted, for each sensor, showing recordings from the first half of the first sessions, and the last half of the last sessions only, for a comparison. The results of this analysis are shown in Figure 10 here below.

The results displayed in Figure 10, show that the novice deploys excessive grip force on sensor 5 (Figure 10, top) of the middle finger, which plays an important role in generating total grip force magnitude for lifting a heavy object with the hand, but is not useful to subtle grip force control of the robotic system. At the end of the last session, the novices grip force is still about five times that of the expert, who deploys only a minimal amount of grip force on sensor 5. The forces deployed by the expert on sensor 6 positioned on the ring finger also evolve differently from the beginning to the end of the sessions compared with those deployed by the novice on that sensor (Figure 10, upper middle). The ring finger contributes largely to grip force control, much less to total force magnitude. The expert deploys stronger grip force on this specific sensor location at the beginning of the sessions, when adjusting the tool and getting in the swing of the control process, considerably less at the end of the sessions. The novice, on the other hand, does not use the ring finger in an noticeably differentiated manner: the grip forces deployed on sensor 6 at the beginning of the sessions are about the same as towards the end of the last session. The largest difference in grip force strategy between the expert and the novice is reflected by the grip force profiles of sensor 7, positioned on the small (pinky) finger of the dominant hand (Figure 10, lower middle). While the expert deploys grip force consistently on that sensor across all sessions from the beginning to the end, within a moderately narrow range of variability for subtle grip force control while steering the robotic handles, the novice hardly deploys any grip force at all on that sensor throughout all sessions. Towards the end, he still shows no signs of expertise in subtle grip force control through strategically deployed force variations in the small finger.

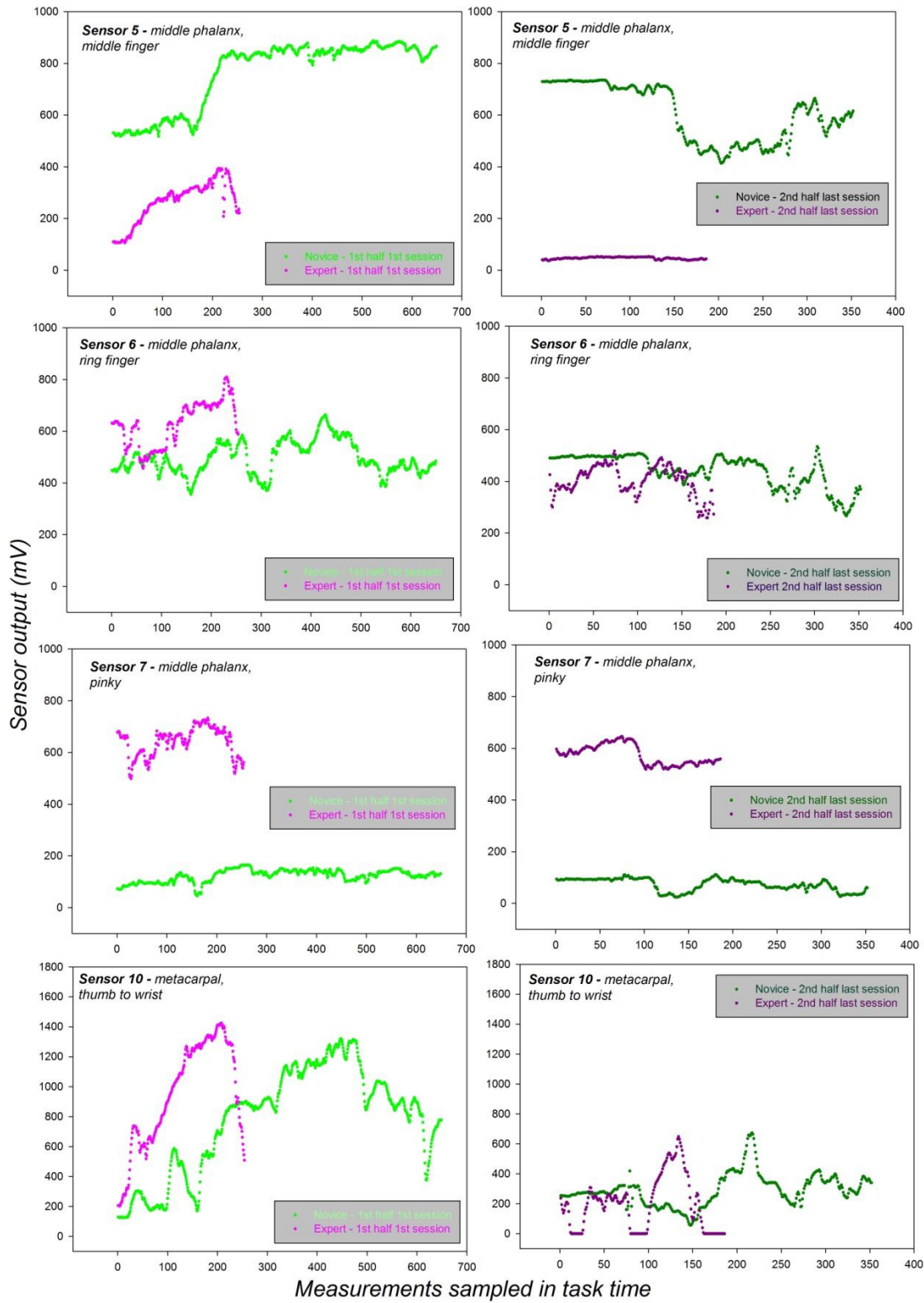

**Figure 10.** Grip force profiles for each of the four specific sensor locations from the first half of the first task sessions, and from the last half of the last task sessions of the expert and the novice performing the task on the robotic system with their dominant hands.

Finally, the grip force profiles that produced the least distinctive characteristics between the expert and the novice are those from sensor 10, positioned on the base metacarpal between thumb and wrist (Figure 10, bottom), a hand region that plays no major role in subtle grip force control, but is important for pushing heavy objects.

### 3.4. Task times and quantitative/qualitative analysis of the task videos

In an additional analysis, the task times for each individual session of the expert and the absolute novice were compared and submitted to analysis of variance. The video sequences captured by the endoscopic camera attached at the distal side of the robotic endoscope and filming the expert and the novice performing the *four step pick-and-drop task* with their dominant hands across the full sequence of individual sessions were analyzed subsequently. The times taken by the expert and the novice in each session to accomplish the robotic system task with the dominant hand are shown in Figure 11.

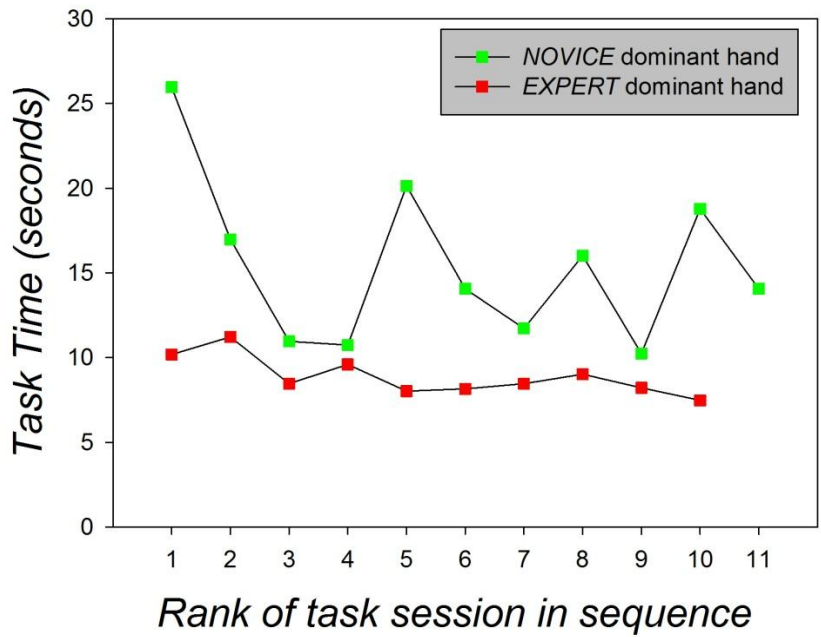

**Figure 11.** Task times from the sessions of the expert and the novice performing the task on the robotic system with their dominant hands.

As shown in Figure 11, the task times of the expert are systematically, often considerably, shorter than the task time of the novice and display very little variance, which is characteristic of highly proficient operators in general. The expert's task times vary between seven and eleven seconds, while those of the novice vary markedly between eleven and twenty seven seconds. The expert's task time in the last session is at the minimum of seven seconds, while the novice takes twice as long (14 seconds) in the last task session with his dominant hand. The difference in task times between the expert was found to be statistically significant on the grounds of statistical paired comparison (Student's t-test). The results are displayed here below in Table 5.

**Table 5.** Paired comparison statistics (Student's t) for the task times of the expert and the novice from ten and eleven successive task sessions respectively

| Group Name | N | Missing | Mean | Stand Dev | SEM |
|---|---|---|---|---|---|
| Time Novice | 11 | 0 | 15.424 | 4.832 | 1.457 |
| Time Expert | 10 | 0 | 8,882 | 1.141 | 0.361 |

**Difference between means:** 6.542

**t = 4.168** with 19 degrees of freedom; **p = <0,001**.

**95 percent confidence interval for difference of means:** 3.257 to 9.827

In a final analysis, quantitative and qualitative criteria for task precision distinguishing the dominant-hand task performance of the expert from that of the novice in the four-step pick-and-drop task were determined. This was possible on the basis of analyses of the video data relative to the individual task sequences filmed by the endoscopic camera of the robotic system. Copies of the original videos from which these analyses were drawn are made available in S2 of the supplementary materials section. The results of these analyses are shown here below in Table 6.

**Table 6.** Quantitative performance analysis relative to task precision of the expert and the novice

|  | *Expert* | Novice |
|---|---|---|
| Tool trajectories towards target requiring adjustment | 3 | 20 |
| Number of times object was accidentally released | 0 | 1 |
| Unsuccessful attempts to grasp object | 1 | 8 |
| Tool collisions with task-space boundary | 0 | 2 |
| Number of times object was dropped outside target area | 0 | 1 |

The results from Table 6 show clear quantitative differences in task precision between the expert and the novice, who has to adjust the tool trajectories many more times in all sessions, and scores a much higher number of unsuccessful grasp attempts compared with the expert, who delivers a close to optimal precision performance. In addition to these quantitative differences, it was noted that the expert adjusted tool movements only very slightly at the beginning of a trajectory. This happened no more than three times in the entire sequence of all eleven task sessions. The novice adjusted tool movements mostly at the end of trajectories, often by multiple adjustments, which happened 20 times in the sequence of his ten task sessions with the dominant hand.

## 4. Discussion

Sensor specific grip force profiles of an expert and an absolute beginner allow for a precise, quantitative and qualitative, characterization of expert control strategies in dominant-hand manipulation of the robotic device handles. Expert force control is characterized here by a marked differential middle and small finger grip force strategy, subtle force modulation using the ring finger and parsimonious grip force deployment on the base metacarpal that joins the thumb to the wrist. The novice uses inappropriate grip forces in these strategic finger-hand regions, producing either excessive or insufficient force at the critical sensor locations. The grip force profiles are consistent with anatomical data relative to the evolution of strategic use of functionally specific finger and hand regions for grip and grip force control in manually executed precision tasks [17-20]. Preconceptions that grip forces may be universally stronger in the dominant hand [18] require a reconsideration in the light of the grip force profiles from this study, which show that the level of expertise and task- device specific factors determine the total amount of grip force deployed by either the dominant or the non-dominant hand. The robotic system exploited here is designed without force feedback, and maneuver control is therefore based solely on visual feedback from the 2D screen images generated by the endoscopic fisheye camera. Different camera system produce image solutions of varying quality, and image quality affects image-guided performance [21, 22, 23]. It will be useful in the future to exploit individual grip force profiling in the context of comparisons between different camera systems for robot-assisted surgery systems. Finally, the sensor glove system developed for this study here will need to be improved for further investigation on larger sample populations. This will allow us to refine expertise benchmarks, and to produce increasingly objective performance criteria for monitoring surgical skill evolution during the training of surgeons on STRAS. The new glove device will ensure that sensor positions can be adapted with high precision to the left and right hands of any group of surgeons or surgical trainees, including women surgeons, whose hands are statistically smaller, as discussed previously in systematic anthropometric studies [20].

**Supplementary Materials:** The following are available online at www.mdpi.com/xxx/s1, **Table S1:** Grip force data in terms of means ($M_{1-10}$) and their standard errors (*SEM*) from sensor-specific individual two-way analyses of variance for ten successive individual task sessions , **Videos S2:** Video sequences captured by the endoscopic camera filming the expert and the novice performing the *four step pick-and-drop task* with their dominant hands.

**Author Contributions:** Conceptualization, B.D.L., F.N., M. de M.; methodology, B.D.L., F.N.; software, P.Z.; validation, F.N., P.Z., B.D.L.; formal analysis, B.D.L., M. de M.; investigation, F.N., B.D.L.; resources, F.N., P.Z.; data curation, F.N.; writing—original draft preparation, B.D.L., F.N., P.Z., M. de M.; writing—review and editing, B.D.L., F.N., M. de M.; visualization, F.N., P.Z.; internal funding acquisition, M. de M., B.D.L., F.N.

**Funding:** This research received no external funding.

**Acknowledgments:** L. Zorn participated in the design of STRAS; A.U. Batmaz, and M.A. Falek provided technical assistance with the sensor glove design, and helped collecting data at an earlier stage of this project. Their respective contributions are gratefully acknowledged.

**Conflicts of Interest:** The authors declare no conflict of interest.

## References

1. C C Thompson, M Ryou, N J Soper, E S Hungess, R I Rothstein, and L L Swanstrom, Evaluation of a manually driven, multitasking platform for complex endoluminal and natural orifice transluminal endoscopic surgery applications (with video), *Gastrointest Endosc*, **2009**, 70 (1), 121–125.

2. L L Swanstrom, R Kozarek, P J Pasricha, S Gross, D Birkett, P O Park, V Saadat, R Ewers, and P Swain, Development of a new access device for transgastric surgery, *J Gastrointest Surg*, **2005**, 9(8), 1129–1137.

3. S J Phee, S C Low, V A Huynh, A P Kencana, Z L Sun, and K Yang, Master and slave transluminal endoscopic robot (MASTER) for natural orifice transluminal endoscopic surgery (NOTES), *Proceedings of the 31st Annual International Conference of the IEEE Engineering in Medicine and Biology Society: Engineering the Future of Biomedicine*, **2009**, 1192–1195.

4. L Zorn, F Nageotte, P Zanne, A. Legner, B. Dallemagne, J. Marescaux, M de Mathelin, A novel telemanipulated robotic assistant for surgical endoscopy: preclinical application to ESD, *IEEE Transactions on Biomedical Engineering*, **2018**, 65(4), 797-808.

5. B. Dallemagne, J. Marescaux, The ANUBIS™ Project, Minimally Invasive Therapy and Allied Technology, **2010**, 19(5), 257-261.

6. A de Donno, F Nageotte, P Zanne, L Zorn, and M de Mathelin, Master/slave control of flexible instruments for minimally invasive surgery, *Proceedings of the IEEE/RSJ International Conference on Intelligent Robots and Systems*, **2013**, 483–489.

7. W R Chitwood, L W Nifong, W H Chapman, J E Felger, B M Bailey, T Ballint, K G Mendelson, V B Kim, J Young, and R Albrecht, Robotic surgical training in an academic institution, *Ann Surg*, **2001**, 234(4), 475-84.

8. C H King, M O Culjat, M L Franco, C E Lewis, E P Dutson, W S Grundfest, and J W Bisley, Tactile feedback induces reduced grasping force in robot-assisted surgery, *IEEE Trans. Haptics*, **2009**, 2(2), 103–110.

9. A Farnè and E Làdavas, Dynamic size-change of hand peripersonal space following tool use, *Neuroreport*, **2000**, 11(8), 1645–1649.

10. A Maravita, M Husain, K Clarke, and J Driver, Reaching with a tool extends visual–tactile interactions into far space: evidence from cross-modal extinction, *Neuropsychologia*, **2001**, 39(6), 580–585.

11. M R Wilson, J S McGrath, S J Vine, J Brewer, D Defriend, and R S W Masters, Perceptual impairment and psychomotor control in virtual laparoscopic surgery, *Surg Endosc Interv Tech*, 25(7), **2011**, 2268–2274.

12. F Sarlegna, J Blouin, J P Bresciani, C Bourdin, J L Vercher, and G M Gauthier, Target and hand position information in the online control of goal-directed arm movements, *Exp Brain Res*, **2003**, 151(4), 524–535.

13. R Pereira, A H J Moreira, M Leite, P L Rodrigues, S Queirós, N F Rodrigues, P Leão, and J L Vilaça, Hand-held robotic device for laparoscopic surgery and training, *Proceedings of the IEEE International Conference on Serious Games and Applications for Health*, **2015**, 1-8.

14. A Legner, M Diana, P Halvax, Y Liu, L Zorn, P Zanne, F Nageotte, M de Mathelin, B. Dallemagne, J Marescaux, Endoluminal surgical triangulation 2.0: A new flexible surgical robot. Preliminary pre-clinical results with colonic submucosal dissection, *Int J Med Robot*, **2017**, 13(3), e1819.

15. P Mascagni, S G Lim, C Fiorillo, P Zanne, F Nageotte, L Zorn, S Peretta, M de Mathelin, J Marescaux, B Dallemagne, Democratizing endoscopic submucosal dissection: single-operator fully robotic colorectal ESD in a pig model. *Gastroenterology*, **2019**, 156(6), 1559-1571.

16. A U Batmaz, A M Falek, L Zorn, F Nageotte, P Zanne, M de Mathelin, B Dresp-Langley, Novice and expert behavior while using a robot controlled surgery system, *IEEE Proceedings of BioMed2017*, Innsbruck, Austria, **2017**, 94-99.

17. S M Cha, H D Shin, K C Kim, and J W Park, Comparison of grip strength among six grip methods, *The Journal of Hand Surgery*, **2014,** 39 (11), 2277-2284.

18. R W Bohannon, Grip strength: a summary of studies comparing dominant and non-dominant limb measurements. *Percept Mot Skills*, **2003**, 96(3), 728-30.

19. R W Young, Evolution of the human hand: the role of throwing and clubbing, *J Anat*, **2003**, 202, 165–174.

20. A G González, D R Rodríguez, J García Sanz-Calcedo, Ergonomic analysis of the dimension of a precision tool handle: a case study, *Procedia Manufacturing*, **2017**, 13, 1336-1343.

21. A U Batmaz, A U, M de Mathelin, B Dresp-Langley, Seeing virtual while acting real: Visual display and strategy effects on the time and precision of eye-hand coordination. *PLoS ONE*, **2017**, *12*(8). https://doi.org/10.1371/journal.pone.0183789

22. B Dresp-Langley, Principles of perceptual grouping: Implications for image-guided surgery. *Front. Psychol.* **2015**, *6*, 1565.

23. J M Wandeto, B Dresp-Langley, The quantization error in a Self-Organizing Map as a contrast and color specific indicator of single-pixel change in large random patterns. *Neural Netw.* **2019**, 119, 273-285.